# LUNG-KGMM: Knowledge-Guided Multimodal Learning for Lung Cancer Incidence Prediction

Chunlei Yang[1], Shuyan Li[2], and Zhong Cao[3,*]

[1] Chinese Academy of Medical Sciences & Peking Union Medical College, Beijing, China

[2] Tsinghua University, Beijing, China

[3] Heidelberg University, Heidelberg, Germany

[*] Corresponding author: Zhong Cao, zhong.cao@uni-heidelberg.de

**Abstract.** Early identification of lung cancer risk is critical for timely intervention, yet existing prediction models are limited by their reliance on single data modalities and their inability to leverage structured clinical knowledge. We propose LUNG-KGMM, a knowledge-guided multimodal framework that integrates longitudinal electronic health records, radiology reports, chest radiograph representations, and guideline-derived knowledge for 1-to-6-year incident lung cancer prediction. To address modality heterogeneity and potential data leakage, we develop a leakage-sanitized report processing pipeline and a horizon-masked cumulative training objective that handles incomplete follow-up. We further introduce a knowledge-graph representation of clinical guidance that encodes report-triggered finding-attribute-action relations as an auditable knowledge stream. We build a multimodal development cohort from the publicly available MIMIC databases and construct a real-world validation cohort from the Xiamen Medical Big Data Platform. Extensive experiments on the MIMIC cohort demonstrate that LUNG-KGMM achieves superior performance over state-of-the-art methods, and validation on the Xiamen cohort further characterizes its cross-cohort portability and the need for local adaptation. The MIMIC development cohort is publicly accessible; the Xiamen cohort is governed by local data privacy regulations.

**Keywords:** lung cancer risk prediction; multimodal learning; electronic health records

## 1 Introduction

Lung cancer remains a leading cause of cancer mortality worldwide [1]. In China, national cancer statistics show that lung cancer remains a major population-health challenge [2], and global cost projections emphasize the macroeconomic burden of cancer [3]. Efficient health-resource allocation is also methodologically important for data-driven health systems [4]. Respiratory comorbidity and smoking-related risk are common in China, as documented by the China Pulmonary Health study and national small-airway dysfunction research [5,6]. Recent Global Burden of Disease analyses further quantify the worldwide burden of chronic respiratory diseases and COPD [7,8]. These clinical and epidemiological realities make early identification of incident lung cancer an important target for routine-care risk modeling.

LDCT screening reduces lung cancer mortality in high-risk populations [9,10]. Screening recommendation statements and clinical guidelines define eligibility, surveillance and program implementation [11-13]. Risk-prediction models and pulmonary-nodule malignancy models further refine CT-screening selection [14,15]. Pulmonary nodule management guidance and reporting systems support follow-up after imaging findings are detected [16,17]. In the present study, chest radiographs are used as routine-care image representations, not as LDCT screening examinations; guideline-derived features provide structured context for report-documented concepts and do not apply LDCT management thresholds to chest-radiograph pixels. Routine-care risk identification is not the same as protocolized screening. Smoking exposure, nodule annotations, follow-up decisions and imaging pixels may be incomplete; radiology reports contain both legitimate index-time risk findings and possible diagnostic-history leakage; and local hospital cohorts may differ from public datasets in language, coding practice and data governance. Deep imaging and multimodal foundation-model studies have shown that future lung cancer risk can be predicted from medical images and multimodal tasks [18-20], but clinical researchers still need a reproducible framework that defines the cohort, aligns modalities, masks unobserved follow-up horizons, tests modality contribution and validates performance in real-world hospital data.

To address these gaps, we developed LUNG-KGMM, as shown in Figure 1, a knowledge-guided multimodal longitudinal learning framework for 1-6-year incident lung cancer prediction. The framework constructs index-time samples, excludes prior lung cancer, creates yearly cumulative labels and follow-up masks, processes leakage-sanitized radiology reports, encodes EHR/report/guideline terms with Transformer sequence encoders, represents chest radiographs with multiview TorchXRayVision features, and uses report-triggered guideline relations as an auditable structured-knowledge stream [21,22]. We used MIMIC-IV to build the public development dataset [23]. MIMIC-CXR and MIMIC-IV Note supplied linked radiographs, reports and clinical notes [24,25], and the Xiamen Medical Big Data Platform was used to construct a large real-world validation cohort. Report-conditioned image generation was evaluated only as a negative missing-image sensitivity analysis and not as a substitute for real imaging evidence [26]. The experimental design first quantifies the contribution of EHR, report, image and guideline streams, then selects image encoding, guideline representation and fusion strategy, and finally compares the selected model with classic and published multimodal baselines under the same 1-6-year incidence endpoints [27,28]. This design allows the study to answer three practical questions: whether a transparent multimodal model can be trained for longitudinal lung cancer incidence prediction, when imaging and structured guideline relations add value beyond strong radiology report representations, and what limitations arise when the same task schema is evaluated across cohorts.

To sum up, our contributions are as follows. First, we construct a large-scale real-world Xiamen lung cancer risk cohort with EHR, radiology reports, and 1-6-year incident lung cancer labels. Second, we propose a KG-guided multimodal cumulative-risk model that integrates EHR, leakage-sanitized radiology reports, CXR, and report-triggered guideline relations, achieving the strongest point estimate among the compared MIMIC models. Finally, we conduct within-cohort evaluation and cross-cohort no-retraining analyses to characterize both performance and portability limitations across MIMIC and Xiamen.

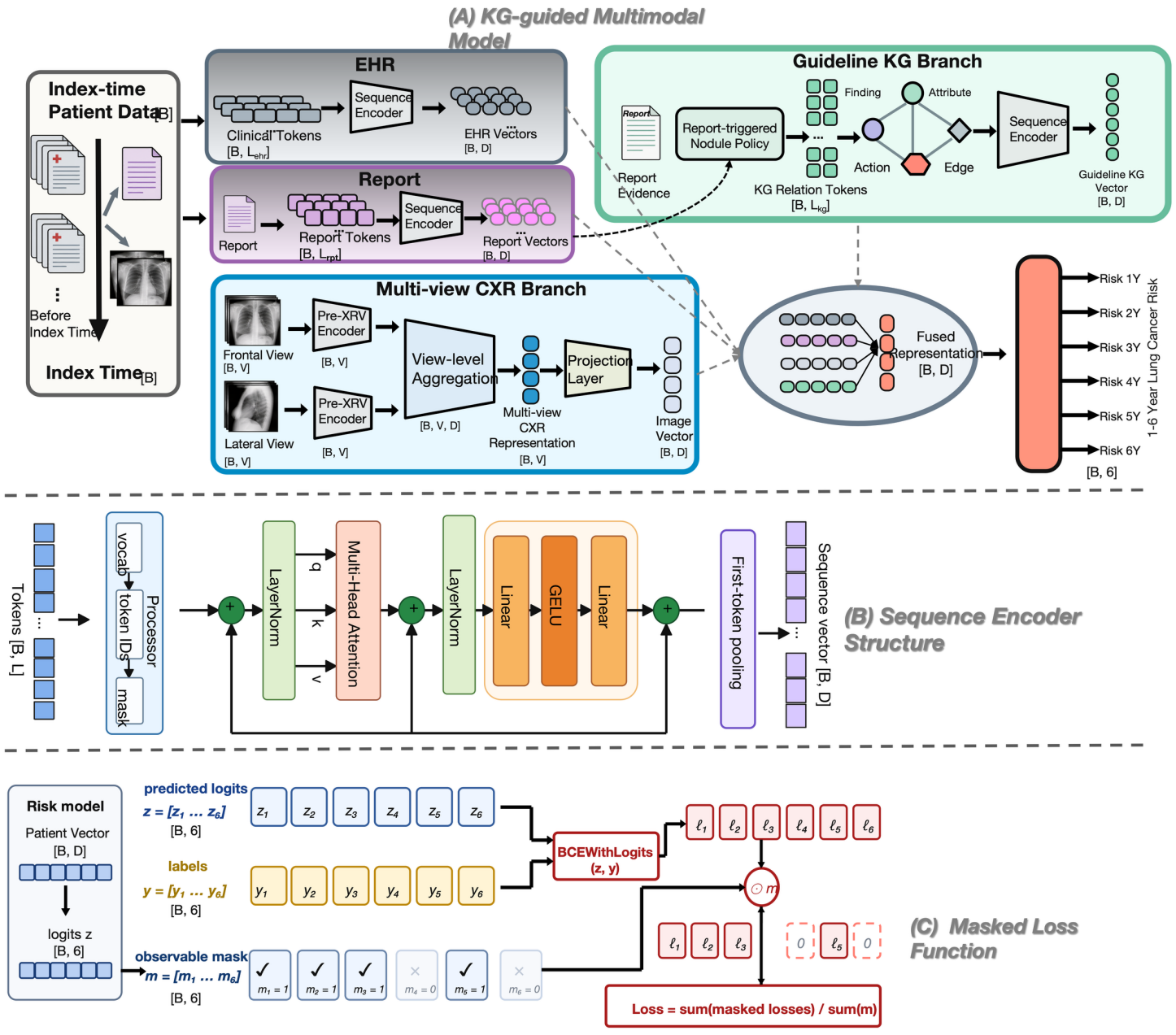

**Fig. 1.** LUNG-KGMM architecture and masked cumulative-risk training. a, Index-time EHR, leakage-sanitized report, report-triggered guideline KG relation tokens and multiview CXR features are encoded into modality vectors and concatenated for 1-6-year risk prediction. KG tokens structure concepts documented in the report and do not apply LDCT management criteria to CXR pixels. b, Sequence encoder: token processors map tokens to IDs and masks; trainable embeddings are passed through Transformer attention, residual normalization and feed-forward layers, followed by first-token pooling. c, Horizon-masked loss applies BCEWithLogitsLoss only to observable cumulative-risk horizons.

## 2 Related Work

EHR-based prediction models provide a complementary route for using routinely collected clinical data in lung cancer risk prediction. Prior neural EHR studies have shown that diagnoses, procedures, medications, laboratory values and longitudinal visit sequences can be transformed into patient-level risk representations through recurrent, attention-based and Transformer-based models, including Doctor AI, RETAIN and Med-BERT [29-31]. Feature-selection and representation-learning studies such as GRASP and unsupervised tractive momentum further illustrate complementary strategies for compact patient-level or low-label feature construction [32,33]. Public datasets such as MIMIC-IV, MIMIC-CXR and MIMIC-IV-Note support reproducible evaluation and multimodal linkage across structured EHR, clinical notes, radiographs and reports [23-25]. However, lung cancer incidence prediction still requires careful definition of index time, prior-cancer exclusion, modality alignment, follow-up masking and external validation, even when using open-source clinical deep learning interfaces such as PyHealth [21].

Medical imaging AI has similarly progressed from single-task chest radiograph classifiers toward longitudinal risk prediction and multimodal clinical modeling. TorchXRayVision established reusable pretrained thoracic imaging representations [22]. Chest radiograph datasets and pretrained encoders such as CheXpert and self-supervised CXR models further support scalable radiographic feature learning [34,35]. LDCT models including Sybil and M3FM demonstrated future lung cancer risk prediction from imaging pixels and multimodal multitask learning [18-20]. Recent multimodal approaches such as MedFuse and DrFuse show the value of combining EHR or clinical time-series data with medical image while handling asynchronous data, missing modalities and modal inconsistency [27,28]. General multimodal biomedical AI studies further motivate structured fusion across heterogeneous medical data streams [36,37]. Guideline knowledge offers an additional structured signal: lung cancer screening and pulmonary nodule guidance encode clinically meaningful relationships among findings, risk modifiers and management actions [12,13]. Pulmonary nodule reporting and management systems further formalize concepts such as nodule size, subsolid appearance, suspicious morphology and follow-up interval [16,17]. Recent guideline-grounded retrieval studies also support using external clinical evidence as an auditable knowledge source for medical AI [38]. Prior studies have shown that pulmonary-nodule findings, attributes and follow-up recommendations can be represented as guideline-derived knowledge-graph relations, and that path-based medical knowledge graphs can augment clinical risk prediction while providing explicit and traceable explanations [39,40].

## 3 Proposed Method

### 3.1 Index-time Multimodal Sample Construction

Each index-time sample used a common schema containing patient identifier, index time, cumulative 1-6-year labels and follow-up masks, pre-index EHR tokens, index radiology report tokens, optional image embeddings and guideline-derived knowledge tokens. EHR variables were restricted to pre-index data and included demographics, diagnoses or conditions, procedures, medications, utilization, vital signs, laboratory tests, smoking-related indicators and clinical note terms when available. Missing fields used field-specific missing tokens. Training-set vocabularies converted token strings into IDs and were reused for validation, test and external-validation splits to avoid vocabulary leakage.

Radiology reports were treated as index-time clinical interpretations. MIMIC reports used alphanumeric term extraction, whereas Xiamen reports used Chinese finding, recommendation, size and report-derived risk terms. Main report-containing experiments used leakage-sensitive sanitization unless explicitly labeled as original: direct cancer diagnosis, prior cancer history and treatment-history terms were removed, while index-time findings such as pulmonary nodule, mass-like lesion, ground-glass opacity, suspicious morphology, size measurement and follow-up recommendation were retained. Report-token length was capped at 256.

### 3.2 Image Representation and Multiview Aggregation

MIMIC-CXR provides real chest radiograph pixels. The default image baseline used CNN layer with grayscale 224-pixel chest radiograph inputs. The main image branch used TorchXRayVision DenseNet121 pretrained chest radiograph features [22]. View-level XRV features were extracted as 1024-dimensional vectors. For studies with multiple available views, frontal PA/AP and lateral embeddings were averaged into a study-level multiview XRV representation. The study-level vector was projected through layer normalization, a linear layer, ReLU activation and dropout before fusion with non-image modalities. For Xiamen, original imaging pixels were not used because of privacy and data-governance restrictions. To explore whether a report-conditioned image branch could partially substitute for missing images, we generated synthetic 2D chest radiograph representations from radiology-report-derived prompts using RoentGen-v2 [26]. These generated images were not treated as real imaging evidence and were not interpreted as LDCT or diagnostic chest radiograph pixels. XRV features were extracted from generated images using the same feature extractor and used only in the Xiamen synthetic-image subset experiments.

### 3.3 Guideline-derived Knowledge

Guideline knowledge was constructed from lung cancer screening guidelines and position statements [12,13]. Pulmonary nodule management guidance and reporting systems supplied a controlled vocabulary of findings, morphology and follow-up concepts [16,17]. These sources were not implemented as executable LDCT management rules. Instead, they provided relational context for concepts explicitly documented in the index radiology report after leakage-sensitive sanitization; the model did not infer nodule size, density or other CT-specific attributes from CXR pixels. Treatment, staging, surgery, chemotherapy, radiotherapy, targeted therapy and immunotherapy content was excluded from primary risk modeling because such information may reflect post-diagnosis management. Trigger concepts included nodule or mass-like lesion, ground-glass or subsolid finding, suspicious morphology, mediastinal or hilar lymph-node abnormality, short-interval follow-up, diagnostic workup recommendation, malignancy-related concern language and explicitly reported size expressions. Size expressions were normalized to millimeters before thresholding. For Xiamen, Chinese report triggers were mapped to corresponding guideline concepts using Chinese keyword rules and a cross-domain canonical vocabulary.

We evaluated several guideline routes, including structured terms plus score, retrieved guideline snippets, executable logic-path tokens and relation-aware knowledge-graph tokens. The final route used knowledge-graph guideline terms. In this route, each documented report trigger was expanded into finding-attribute-action relation tokens. For example, a nodule mention generated nodes and edges representing nodule presence and the guideline relation that size and density are relevant assessment attributes; it did not assert an unreported size or density value. A ground-glass or subsolid mention generated relation tokens for persistence assessment. Suspicious morphology generated relation tokens for spiculation or lobulation increasing risk. Diagnostic workup recommendations generated action tokens indicating higher documented clinical suspicion. Because these triggers originate from the report, the KG is a structured re-encoding of report evidence augmented with guideline relations, rather than an independent clinical observation or an automated management recommendation.

### 3.4 Model Architecture

The model followed a PyHealth-style sample interface with modality-specific encoders and a cumulative-risk prediction head. Sequence features, including EHR fields, radiology report terms and KG guideline terms, were declared as sequence inputs. Sequence processors built vocabularies from the training split and converted token strings into integer IDs; validation, test and external-validation splits reused the same processors to avoid vocabulary leakage. The embedding layer mapped token IDs into 128-dimensional trainable embeddings. Each sequence field was encoded by a one-layer Transformer encoder with two attention heads, dropout 0.2 and maximum sequence length 192 for EHR fields or 256 for report terms. The pooled field-level output was used as the fixed-dimensional representation for that field. EHR field representations were concatenated into an EHR vector, report tokens were encoded into a report vector, and KG guideline terms were encoded into a guideline vector. Image features were projected into the same fusion space through a layer-normalized linear projection block.

The LUNG-KGMM combined EHR tokens, leakage-sanitized report tokens, multiview XRV image features and KG guideline terms using concat fusion. Attention and gated fusion were evaluated under the same complete KG-input setting as ablation experiments, but concat fusion was retained because it achieved the highest macro 1/3/6-year AUROC in the LUNG-KGMM architecture comparison. The output head produced six cumulative risk logits corresponding to 1-, 2-, 3-, 4-, 5- and 6-year lung cancer incidence risk. The cumulative-risk layer predicted horizon-specific incremental components and a base risk term, applied an upper-triangular cumulative transformation to enforce monotonic cumulative logits, and converted the six logits into predicted risks using the sigmoid function.

### 3.5 Training Objective and Optimization

The primary objective was horizon-masked cumulative binary cross-entropy. For each sample, BCEWithLogitsLoss was computed over the six cumulative logits and six cumulative labels, weighted by the horizon-specific follow-up mask and normalized by the number of observable horizons. Unobserved horizons were not treated as negative outcomes. The final objective did not use class positive weights. Earlier positive-weighted, unmasked complete-case, visit-level/time-bin and alternative cumulative-risk variants were used during method selection but were not retained for the main model. Models were trained using the PyHealth Trainer and PyTorch. Training used Adam optimization with learning rate 1e-3, weight decay 1e-5, maximum gradient norm 5.0, dropout 0.2, batch size 128 and evaluation batch size 256 unless otherwise specified. Validation loss was monitored with early stopping patience 4. Training, validation and test datasets shared the training-split input processors and output processors.

# 4 Experiments

## 4.1 Datasets and Settings

This study used a public MIMIC development cohort and a Xiamen Medical Big Data Platform analysis cohort. The cohort summary is shown in Table 1. The MIMIC cohort was built from MIMIC-IV 3.1, MIMIC-IV Note, MIMIC-CXR images and MIMIC-CXR reports [23-25], and was used for the main model development, method comparison and ablation analyses because it contained EHR, radiology reports and CXR images. The Xiamen cohort provided real-world EHR and radiology-report data under privacy restrictions and was used for within-cohort evaluation and no-retraining cross-cohort portability analysis of the same task pipeline. A synthetic-image subset was analyzed only as a negative missing-image sensitivity experiment. Primary metrics were 1- to 6-year AUROC, AUPRC and Brier score, with macro 1/3/6-year AUROC used for model selection. Comparators included LR, MLP, RNN-GRU, RETAIN, Transformer, and task-adapted MedFuse-style and DrFuse-style models under the same lung cancer incidence endpoints.

**Table 1.** Cohort and modality summary.

| Cohort | Samples | Train / Val / Test | Outcome horizons |
|---|---|---|---|
| MIMIC public cohort | 20,626 | 14,438 / 3,094 / 3,094 | 1-6 years |
| Xiamen E-cohort | 127,912 | 89,319 / 19,222 / 19,371 | 1-6 years |
| Xiamen synthetic-image subset | 57,476 | 40,141 / 8,611 / 8,724 | 1-6 years |

## 4.2 MIMIC Cohort Benchmark Comparison

As shown in Table 2, we evaluated LUNG-KGMM against classic model baselines and published EHR-CXR fusion architectures on the same MIMIC cohort, split and 1/3/6-year incidence endpoints. LR, MLP, RNN-GRU, RETAIN and Transformer baselines used the EHR and report inputs supported by their original tabular/text design; MedFuse-style and DrFuse-style baselines supplied EHR and report together with a single-index CXR feature.

The strongest classic baseline was MLP with macro 1/3/6-year AUROC 0.870. MedFuse-style fusion achieved macro AUROC 0.825 and DrFuse-style fusion achieved macro AUROC 0.876. EHR, reports and image features were included where supported by each baseline architecture. The selected LUNG-KGMM model achieved macro AUROC 0.885, macro AUPRC 0.507 and Brier score 0.034. Its absolute macro-AUROC gains were 0.015 over MLP and 0.009 over DrFuse-style fusion. These are modest point-estimate improvements that support incremental model value on this test split but do not by themselves establish clinical benefit, which requires prospective and decision-analytic evaluation.

**Table 2.** MIMIC method-family baseline comparison. All rows use the same MIMIC test split and 1/3/6-year lung cancer incidence endpoints. Report-containing rows use leakage-sanitized reports. LR/MLP/RNN/RETAIN/Transformer rows use the inputs supported by their original tabular/text design. MedFuse-style and DrFuse-style rows use task-adapted EHR-CXR fusion architectures with EHR, leakage-sanitized report and a single-index CXR feature, but without KG guideline terms or multiview aggregation. LUNG-KGMM denotes the selected complete-input model.

| Method | Macro AUROC | Macro AUPRC | Macro Brier |
|---|---|---|---|
| LR | 0.865 | 0.402 | 0.038 |
| MLP | 0.870 | 0.488 | 0.035 |
| RNN-GRU | 0.818 | 0.350 | 0.039 |
| RETAIN | 0.787 | 0.233 | 0.042 |
| Transformer | 0.815 | 0.380 | 0.038 |
| CNN | 0.620 | 0.099 | 0.048 |
| MedFuse-style | 0.825 | 0.319 | 0.040 |
| DrFuse-style | 0.876 | 0.456 | 0.035 |
| **LUNG-KGMM** | **0.885** | **0.507** | **0.034** |

LUNG-KGMM enriched high-risk subgroups. At the top 10% risk threshold, positive predictive value was 0.135 for 1-year risk, 0.271 for 3-year risk and 0.510 for 6-year risk. Sensitivity at the same threshold was 0.764, 0.718 and 0.541. At specificity >=0.95, sensitivity was 0.709, 0.679 and 0.541 for 1-year, 3-year and 6-year risk, respectively, as shown in Table S1.

Figure 2 summarizes discrimination, precision-recall performance and calibration. It shows that LUNG-KGMM maintains stronger ROC and precision-recall performance than representative image-only and tri-modal/no-guideline comparators, while the calibration plots show agreement between predicted and observed risks. Clinical operating thresholds and high-risk-strata enrichment are summarized separately in Table S1.

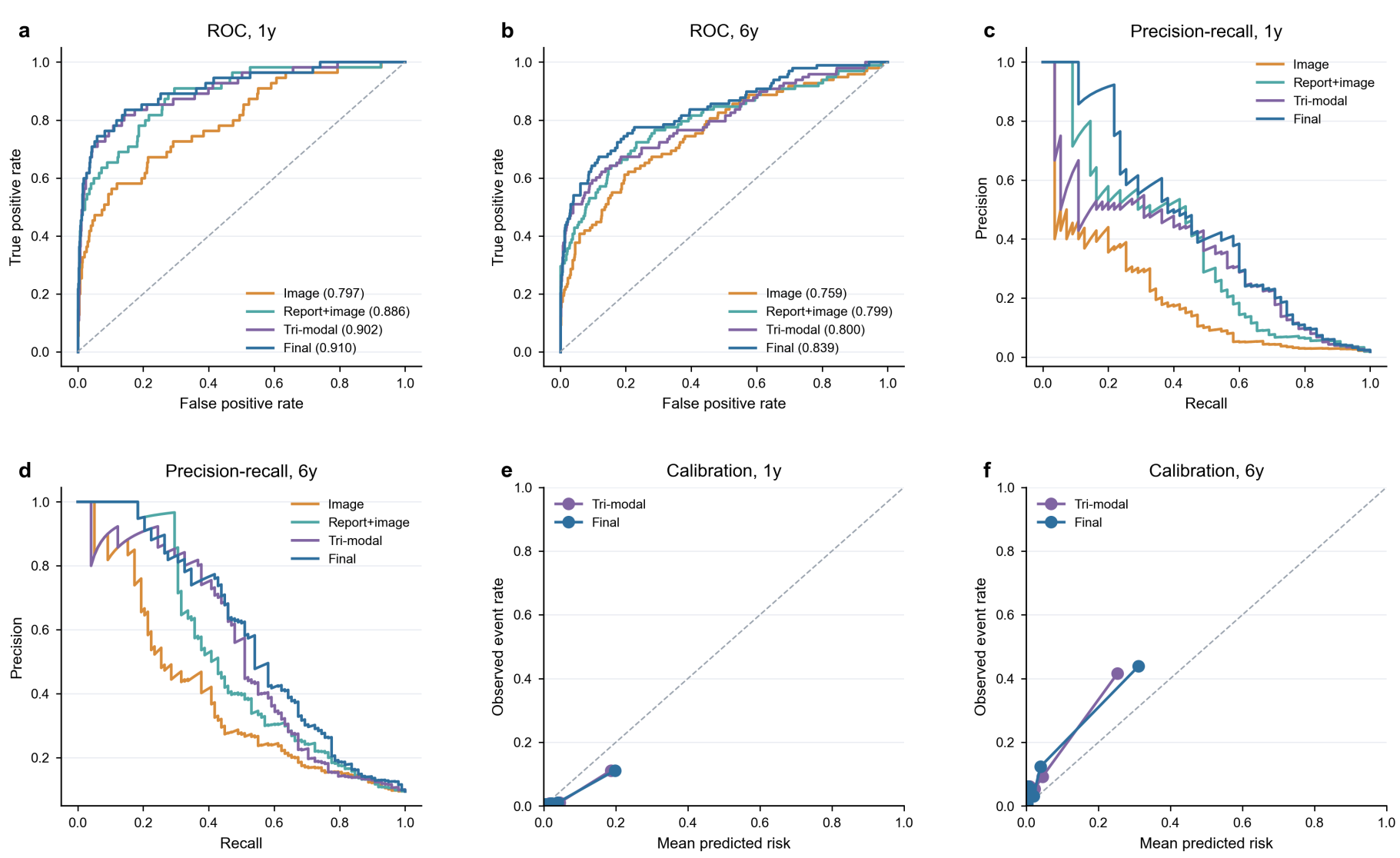

**Fig. 2.** MIMIC test-set discrimination, precision-recall and calibration curves. a-b, ROC curves for 1-year and 6-year prediction. c-d, Precision-recall curves. e-f, Calibration curves for representative multimodal models.

### 4.3 Ablation Studies

**Single-modality.** As shown in Table 3, EHR-only model achieved macro 1/3/6-year AUROC 0.743, AUPRC 0.163 and Brier score 0.046. Report-only model was substantially stronger, with macro AUROC 0.854 using original reports and 0.873 after leakage-sensitive report sanitization. The CNN image branch reached macro AUROC 0.620, whereas TorchXRayVision single-view image features reached 0.763 and multiview XRV features reached 0.787.

These findings indicate that radiology reports carry strong semantic risk information. This is clinically expected because a report is a radiologist-produced compression of image findings, including nodules, mass-like lesions, ground-glass components, suspicious morphology and follow-up recommendations. Leakage sanitization removed explicit cancer diagnosis and treatment-history language, but it cannot remove all diagnostic-concern wording without also deleting legitimate index-time risk findings. The report-only AUROC of 0.873 should therefore be interpreted as a strong semantic baseline that may retain diagnostic-proximity signal, while image and KG contributions are incremental. Image-only performance was lower than report-only performance, but domain-specific pretrained image features substantially improved over the CNN image branch.

**Guideline representation.** Guideline knowledge was introduced as a clinically traceable feature set. Structured guideline terms/scores, retrieved guideline text, executable logic-path tokens and knowledge-graph relation terms were evaluated under the same leakage-sanitized report setting. The knowledge-graph route encoded guideline concepts as relation-aware terms connecting findings, risk modifiers and management concepts, and provided the best guideline representation in MIMIC.

To isolate guideline representation from fusion strategy, all guideline routes in this ablation were evaluated under the same concat cumulative-risk architecture, with EHR, leakage-sanitized report and multiview XRV streams fixed. As shown in Table 3, compared with the no-guideline model (macro AUROC 0.860), structured terms plus score reached 0.876, retrieved guideline text reached 0.873, executable path plus score reached 0.868, and knowledge-graph terms reached 0.885 with AUPRC 0.507 and Brier score 0.034.

These results suggest that report-triggered guideline information was most useful when represented as relations rather than as long retrieved text. The KG route compactly linked report findings to risk modifiers and management concepts, making the transformation auditable while avoiding a large free-text context. The KG is partially redundant with the report by design and should not be presented as an independent modality; the increase from 0.860 without guideline terms to 0.885 with KG terms supports a useful structured representation on this split, not clinically independent knowledge or causal benefit.

**Fusion strategy.** As shown in Table 3, fusion strategies were compared under a fixed complete LUNG-KGMM input: EHR, leakage-sanitized report, multiview XRV image features and KG guideline terms. Concat fusion achieved the highest macro AUROC 0.885 and AUPRC 0.507, attention fusion reached macro AUROC 0.853 and AUPRC 0.422, and gated fusion reached macro AUROC 0.874 and AUPRC 0.490 while giving the lowest Brier score 0.033. Concat fusion was therefore retained because model selection was based on macro 1/3/6-year AUROC.

**Final input stack.** The best-performing modality combination used EHR, leakage-sanitized report, multiview XRV image features and KG guideline terms. The report stream provided the largest gain over EHR alone, whereas multiview XRV and KG guideline terms added smaller but complementary improvements. The final LUNG-KGMM stack reached macro AUROC 0.885, AUPRC 0.507 and Brier score 0.034.

**Table 3.** Full ablation summary for modality contribution, report leakage sensitivity, modality combination, guideline representation and fusion strategy on MIMIC.

| Ablation block | Variant | Macro 1/3/6 AUROC | Macro 1/3/6 AUPRC | Macro 1/3/6 Brier |
|---|---|---|---|---|
| Single modality | EHR only | 0.743 | 0.163 | 0.046 |
| **Single modality** | **Report only, original** | **0.854** | **0.424** | **0.036** |
| **Single modality** | **Report only, leakage-sanitized** | **0.873** | **0.432** | **0.036** |
| Single modality / Image encoder | CNN image branch | 0.620 | 0.099 | 0.048 |

| Ablation block | Variant | Macro 1/3/6 AUROC | Macro 1/3/6 AUPRC | Macro 1/3/6 Brier |
|---|---|---|---|---|
| Image encoder | XRV image branch, single view | 0.763 | 0.286 | 0.040 |
| **Image encoder** | **XRV image branch, multiview** | **0.787** | **0.298** | **0.040** |
| Modality combination | EHR | 0.743 | 0.163 | 0.046 |
| Modality combination | EHR + report | 0.845 | 0.391 | 0.039 |
| Modality combination | EHR + multiview XRV | 0.820 | 0.357 | 0.041 |
| Modality combination | EHR + report + multiview XRV | 0.860 | 0.484 | 0.034 |
| **Modality combination** | **LUNG-KGMM** | **0.885** | **0.507** | **0.034** |
| Guideline representation | No guideline | 0.860 | 0.484 | 0.034 |
| Guideline representation | Structured terms + score | 0.876 | 0.475 | 0.035 |
| Guideline representation | Retrieved guideline text | 0.873 | 0.499 | 0.034 |
| Guideline representation | Executable path + score | 0.868 | 0.497 | 0.035 |
| **Guideline representation** | **Knowledge-graph terms** | **0.885** | **0.507** | **0.034** |
| **Fusion strategy** | **Concat** | **0.885** | **0.507** | 0.034 |
| Fusion strategy | Attention | 0.853 | 0.422 | 0.036 |
| Fusion strategy | Gated | 0.874 | 0.490 | **0.033** |

*Note: Report-containing rows use leakage-sanitized report tokens unless explicitly labeled as original report. Complete input denotes EHR, leakage-sanitized report, multiview XRV image features and the specified guideline representation.*

### 4.4 Real-world Validation in Xiamen E-cohort

As shown in Table 4, two no-retraining transfer analyses were evaluated. The primary no-image setting used EHR, leakage-sanitized report and KG guideline terms. A separate negative sensitivity setting added XRV features, corresponding to multiview real CXR embeddings in MIMIC and report-conditioned synthetic XRV features in Xiamen. The no-image transfer achieved MIMIC-to-Xiamen macro 1/3/6-year AUROC 0.750 and Xiamen-to-MIMIC AUROC 0.706. Adding the mismatched real-versus-synthetic image stream reduced these values to 0.592 and 0.642, respectively. The image-enabled transfer was weaker, consistent with an image-domain mismatch between real MIMIC CXR embeddings and Xiamen synthetic report-conditioned images. These results show that the task schema can be executed in both cohorts, but the learned weights are not directly transportable across English and Chinese reports or across real and report-conditioned synthetic image representations.

These transfer results characterize partial portability of the task schema and model architecture, not external generalizability of fixed weights. Differences in language, reporting practice, healthcare system and image availability require local terminology harmonization, feature alignment, recalibration and likely site-specific retraining before deployment.

**Table 4.** Cross-cohort external validation without retraining.

| Transfer direction | Input | Target N | Macro 1/3/6 AUROC | Macro 1/3/6 AUPRC | Macro 1/3/6 Brier | Macro 1-6 AUROC |
|---|---|---|---|---|---|---|
| MIMIC -> Xiamen | EHR + report + KG | 19,371 | 0.750 | 0.213 | 0.086 | 0.748 |
| Xiamen -> MIMIC | EHR + report + KG | 3,094 | 0.706 | 0.125 | 0.049 | 0.710 |
| MIMIC -> Xiamen synthetic subset | EHR + report + synthetic XRV + KG | 8,724 | 0.592 | 0.049 | 0.027 | 0.579 |
| Xiamen synthetic subset -> MIMIC | EHR + report + synthetic XRV + KG | 3,094 | 0.642 | 0.113 | 0.053 | 0.648 |

## 5 Limitations

This study has several limitations. First, MIMIC-CXR provides chest radiographs rather than LDCT. The KG structures concepts documented in radiology reports but includes relations derived from CT-oriented guidance; it must not be interpreted as applying Lung-RADS or Fleischner management criteria to CXR pixels or as demonstrating guideline-adherent clinical decisions. Second, the sanitized report-only model reached macro AUROC 0.873, close to the complete model's 0.885. Residual diagnostic-concern language may remain, and the image/KG gains are modest incremental improvements rather than evidence of independent clinical utility. Third, no-retraining cross-cohort performance decreased substantially, indicating that local terminology mapping, recalibration and likely retraining are required. Fourth, report-conditioned synthetic images were a negative sensitivity analysis and should not be treated as substitutes for unavailable real images. Xiamen imaging pixels were not used because of privacy constraints, and smoking and BMI were incompletely captured. Finally, all analyses were retrospective and require prospective validation before clinical deployment.

## 6 Conclusion

LUNG-KGMM provides a reproducible research framework for 1-6-year lung cancer incidence risk prediction. In MIMIC, it integrates longitudinal EHR, leakage-sanitized radiology reports, pretrained chest-radiograph representations and report-triggered guideline relations while preserving explicit cohort construction, follow-up masking and modality ablation. The Xiamen real-world validation demonstrates that the task schema can be evaluated in a data-governed clinical setting, but fixed model weights require local adaptation and prospective validation before clinical use.

**Disclosure of Interests.** The authors have no competing interests to declare that are relevant to the content of this article.

# Supplementary Appendix for PRCV 2026 LUNG-KGMM Submission

## Supplementary Methods

### Datasets and Experiment Setting

This study used both public de-identified research data and a real-world hospital-derived cohort. The public development cohort was constructed from MIMIC-IV 3.1 and incorporated MIMIC-IV Note, MIMIC-CXR chest radiographs and MIMIC-CXR radiology reports, which are available through credentialing and data-use agreements. As shown in Figure S1, the Xiamen cohort was extracted from the Xiamen health-data platform under local institutional approval and data-security review. It contained structured EHR data, clinical note-derived information and radiology reports; original imaging pixels existed in the source environment but were not used in this project because of privacy and data-access restrictions.

All experiments used index-time task construction. Records before the index time contributed historical EHR features, the index radiology report contributed report tokens, and outcomes were defined as cumulative lung-cancer incidence over 1-6 years. Horizon-specific follow-up masks ensured that unobservable horizons were excluded rather than counted as negative outcomes. Model selection used macro 1/3/6-year AUROC, with macro AUPRC and Brier score reported as secondary metrics.

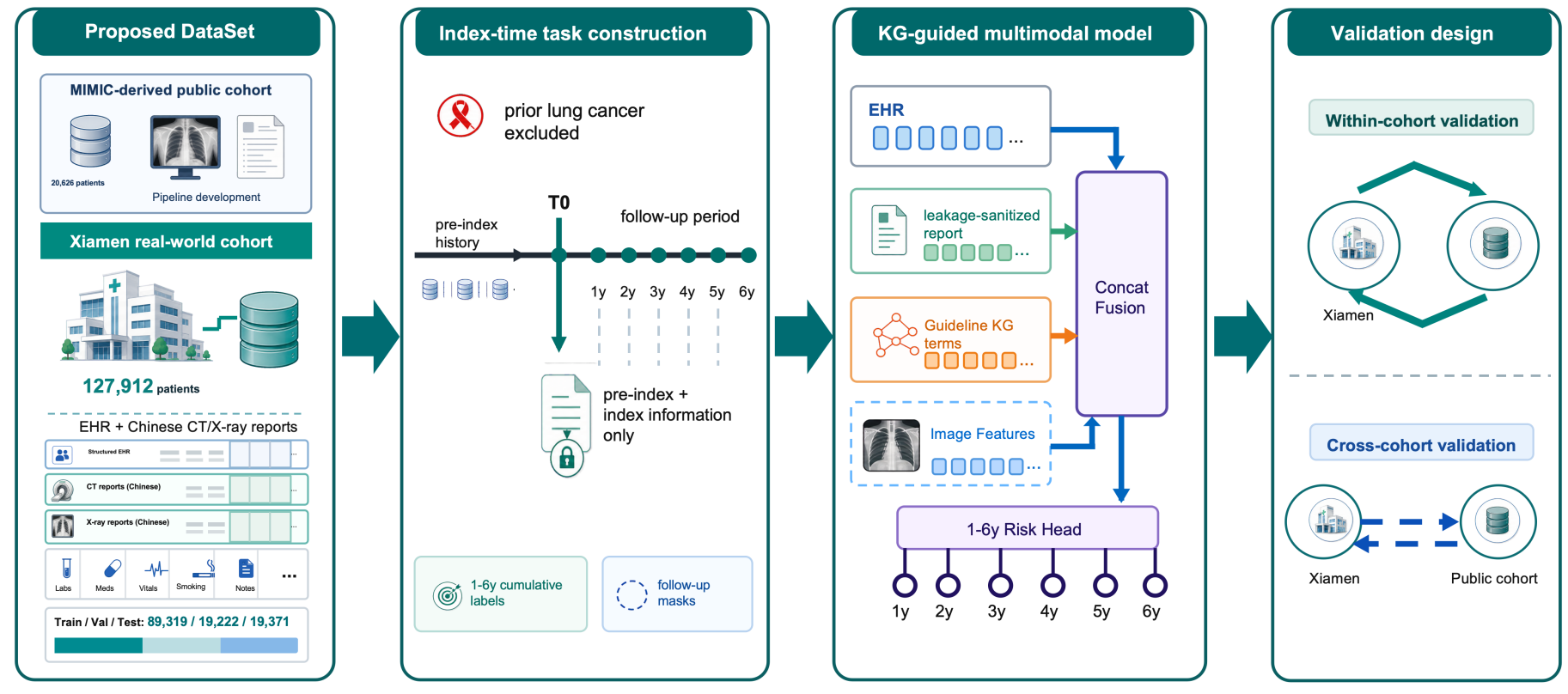

**Fig. S1. Study overview and validation design. LUNG-KGMM constructs index-time 1-6-year lung cancer incidence tasks from Xiamen and public cohorts, fuses EHR, leakage-sanitized report, guideline KG and available image streams, and evaluates within-cohort performance, cross-cohort portability and an exploratory synthetic-image sensitivity analysis.**

## Within-cohort, Synthetic-image and Cross-cohort Experiments

Within-cohort experiments trained, selected and tested models within the same cohort using patient-level train/validation/test splits. In MIMIC, complete multimodal experiments evaluated EHR, leakage-sanitized report, real CXR-derived XRV features and KG guideline terms. In Xiamen, the primary full-cohort experiments evaluated EHR, leakage-sanitized report and KG guideline terms because original imaging pixels were not used in this project. A separate Xiamen synthetic-image subset used report-conditioned synthetic CXR representations to examine whether an image-like branch could partially substitute for missing original imaging pixels.

Strict cross-cohort no-retraining analysis was used only as a portability sensitivity test. Models trained in one cohort were evaluated in the other cohort without retraining. No-image KG transfer used EHR, leakage-sanitized report and KG guideline terms. A negative image-domain sensitivity setting additionally used image_xrv features, corresponding to real CXR-derived XRV features in MIMIC and report-conditioned synthetic XRV features in Xiamen. These experiments assess portability of the task schema and model architecture, not direct transportability of model weights.

## Radiology report processing and leakage-sensitive sanitization

Radiology reports were treated as index-time clinical interpretations of the imaging examination. MIMIC reports were tokenized using alphanumeric term extraction and domain prefixes. Xiamen reports were processed with Chinese term extraction, including finding terms, recommendation terms, size-related expressions and report-derived risk indicators. The maximum report-token length was 256.

Because radiology reports may contain both legitimate index-time risk findings and direct diagnosis or treatment-history terms, all main report-containing experiments used leakage-sensitive report sanitization unless explicitly labeled as original. Sanitization removed direct cancer diagnosis, prior cancer history and treatment-history language, including terms corresponding to lung cancer, malignancy, carcinoma, metastasis, staging, chemotherapy, radiotherapy, lobectomy, pneumonectomy and postoperative treatment history. Index-time risk findings such as pulmonary nodule, mass-like lesion, ground-glass opacity, suspicious morphology, size measurement and follow-up recommendation were retained.

## Literature Baselines

MedFuse-style and DrFuse-style baselines were implemented as task-adapted literature comparators. MedFuse represents a representative EHR-CXR fusion design that combines a clinical-sequence branch and an image branch for multimodal prediction. DrFuse extends this idea with a disentangled multimodal-fusion strategy intended to improve robustness when modalities are partially missing or carry complementary information. In our experiments, both baselines used the same patient-level split, outcome labels, follow-up masks, leakage-sanitized report information, CXR feature input when available, optimizer protocol and evaluation metrics as the main experiments.

These literature baselines did not receive the LUNG-KGMM-specific additions of KG guideline terms or multiview XRV aggregation unless explicitly stated. This design makes the comparison a method-family comparison rather than a test of whether giving every baseline the final model's domain-specific components improves performance.

## Supplementary Results

This supplementary analysis evaluated clinically interpretable operating points for LUNG-KGMM in the public MIMIC test set. We examined two risk-stratification settings: the top 10% highest predicted-risk group and thresholds selected to maintain specificity at or above 0.95. In the top 10% predicted-risk stratum, positive predictive value increased from 0.135 at 1 year to 0.510 at 6 years, indicating that longer-horizon predictions enriched substantially for future lung cancer events. Sensitivity in the same high-risk stratum was 0.764, 0.718 and 0.541 at 1, 3 and 6 years, respectively. At the high-specificity operating point, LUNG-KGMM maintained sensitivities of 0.709, 0.679 and 0.541 across the 1-, 3- and 6-year horizons. These results suggest that the model is most useful as a risk-stratification and prioritization tool, identifying a clinically meaningful high-risk subgroup while preserving high specificity.

**Table S1.** Clinical operating points for LUNG-KGMM in the public MIMIC test set.

| Operating point | 1-year | 3-year | 6-year |
|---|---|---|---|
| Top 10% predicted-risk PPV | 0.135 | 0.271 | 0.510 |
| Top 10% predicted-risk sensitivity | 0.764 | 0.718 | 0.541 |
| Sensitivity at specificity >=0.95 | 0.709 | 0.679 | 0.541 |

**Note:** PPV denotes positive predictive value. Values correspond to the clinical operating-point analysis cited in the main text.

### Xiamen Within-cohort Validation

We evaluated LUNG-KGMM in the Xiamen real-world cohort using within-cohort training, validation and testing. Because original imaging pixels were not used in this project, the primary Xiamen analysis used EHR, leakage-sanitized radiology reports and KG guideline terms. This experiment assessed whether the selected model design remained effective in a large Chinese hospital-based cohort under the available-modality setting.

As shown in Table S2, EHR-only and report-only models achieved similar macro 1/3/6-year AUROC values (0.873 and 0.870, respectively), whereas combining EHR and report improved AUROC to 0.912. Adding KG guideline terms further improved AUROC to 0.919 and AUPRC to 0.683, with a Brier score of 0.038. The gain was smaller than the report stream itself but was consistent with the role of guideline knowledge as a structured, auditable refinement of report-derived risk information.

**Table S2. Xiamen full-cohort within-cohort validation using EHR, reports and KG guideline terms.**

| Model | N | Input | Macro 1/3/6 AUROC | Macro 1/3/6 AUPRC | Macro 1/3/6 Brier | Macro 1-6 AUROC |
|---|---|---|---|---|---|---|
| EHR only | 127,912 | Structured EHR | 0.873 | 0.528 | 0.045 | 0.876 |
| Report only | 127,912 | Report | 0.870 | 0.543 | 0.049 | 0.868 |
| EHR + report | 127,912 | EHR + leakage-sanitized report | 0.912 | 0.672 | 0.037 | 0.908 |
| **LUNG-KGMM no-image** | 127,912 | EHR + leakage-sanitized report + KG guideline terms | **0.919** | **0.683** | **0.038** | **0.917** |

**Note:** The Xiamen full-cohort analyses did not use original imaging pixels in this project because of privacy and data-access restrictions.

Figure S2 shows the corresponding Xiamen within-cohort discrimination, precision-recall and calibration curves. The EHR+report+KG model retained strong ROC performance at both 1-year and 6-year horizons and improved precision-recall performance over single-stream models. Calibration curves are shown for EHR+report and LUNG-KGMM to compare the report-based model with the KG-augmented final Xiamen configuration.

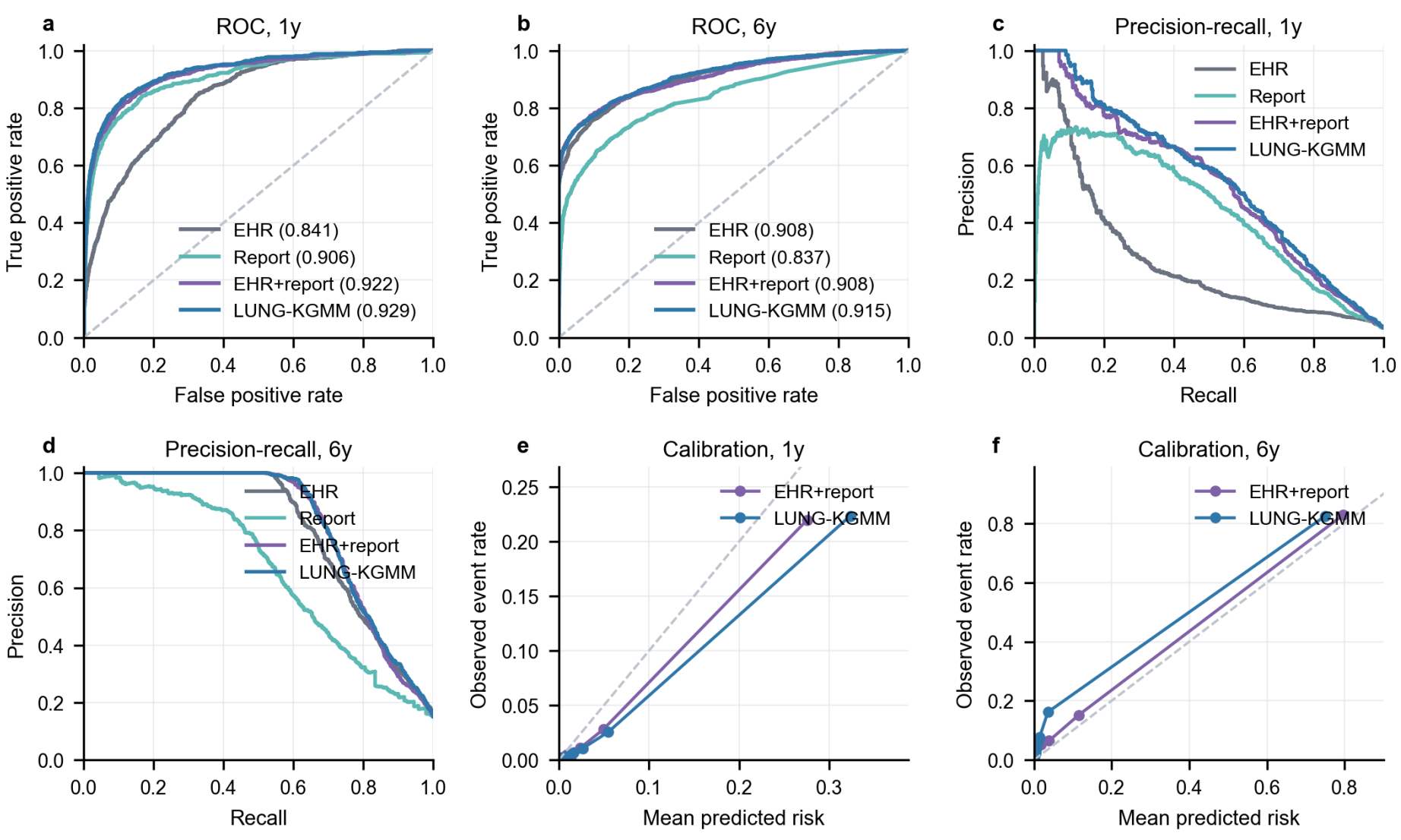

**Figure S2.** Xiamen within-cohort test-set discrimination, precision-recall and calibration curves. a-b, ROC curves for 1-year and 6-year prediction. c-d, Precision-recall curves. e-f, Calibration curves for EHR+report and LUNG-KGMM. All report-containing rows use leakage-sanitized report tokens.

### Xiamen Report-conditioned Synthetic-image Negative-control Analysis

The report-conditioned synthetic-image analysis was retained as an exploratory negative-control missing-image analysis, not as evidence from real imaging pixels. In the 57,476-patient subset with RoentGen-v2 synthetic CXR features, synthetic image features alone showed limited discrimination, whereas EHR plus synthetic image features reached macro 1/3/6-year AUROC 0.810. The leakage-sanitized KG-augmented synthetic multimodal model reached macro AUROC 0.809 and AUPRC 0.219. Non-KG report-containing leakage-sanitized pilot analyses were available on a smaller 1,000-sample subset and are reported as negative sensitivity checks rather than as LUNG-KGMM-selection experiments.

The synthetic-image findings should therefore be interpreted as missing-modality sensitivity analysis rather than evidence that generated images replace real imaging pixels. Synthetic representations carried some signal, but they did not add clear value beyond the original report text in the current Xiamen setting.

**Table S3.** Xiamen synthetic-image subset sensitivity analysis using report-conditioned synthetic image features.

| Model | N | Input | Macro 1/3/6 AUROC | Macro 1/3/6 AUPRC | Macro 1/3/6 Brier |
|---|---|---|---|---|---|
| Synthetic image only | 57,476 | Report-conditioned synthetic XRV features | 0.605 | 0.039 | 0.026 |
| EHR + synthetic image | 57,476 | EHR + synthetic XRV features | 0.810 | 0.198 | 0.024 |
| EHR + report | 1,000 | EHR + report, pilot subset | 0.718 | 0.057 | 0.010 |
| EHR + synthetic image + report | 1,000 | EHR + synthetic XRV + report, pilot subset | 0.694 | 0.054 | 0.010 |
| KG-augmented synthetic multimodal model | 57,476 | EHR + synthetic XRV + report + KG guideline terms | 0.809 | 0.219 | 0.025 |

**Note:** Synthetic images were used only as a missing-image sensitivity branch and should not be interpreted as real imaging evidence.